\documentclass[runningheads]{llncs}
\usepackage[T1]{fontenc}
\usepackage{graphicx}
\usepackage{booktabs}
\usepackage{multirow}
\usepackage{subcaption}
\usepackage{fontawesome5}
\usepackage[dvipsnames,table]{xcolor}
\usepackage{todonotes}
\usepackage{hyperref}

\hypersetup{
    colorlinks=true,
    linkcolor=black,
    filecolor=black,      
    urlcolor=blue,
    citecolor=blue
}
\begin{document}
\title{IXAII: An Interactive Explainable Artificial Intelligence Interface for Decision Support Systems}
\titlerunning{IXAII: An Interactive Explainable AI Interface}
\author{Pauline Speckmann\inst{1}\orcidID{0009-0007-5290-5118} \and
    Mario Nadj\inst{2}\orcidID{0000-0002-6901-9450} \and
    Christian Janiesch\inst{1}\orcidID{0000-0002-8050-123X}
}
\authorrunning{P. Speckmann et al.}
\institute{
    TU Dortmund University, Dortmund, Germany\\
        \email{$\{$pauline.speckmann, christian.janiesch$\}$@tu-dortmund.de}
    \and
    University of Duisburg-Essen, Essen, Germany\\
        \email{mario.nadj@ris.uni-due.de}
}
\maketitle
\begin{abstract}
% Soll: 150--250 words, Ist: 149 words lol
Although several post-hoc methods for explainable AI have been developed, most are static and neglect the user perspective, limiting their effectiveness for the target audience. In response, we developed the interactive explainable intelligent system called IXAII that offers explanations from four explainable AI methods: LIME, SHAP, Anchors, and DiCE. Our prototype provides tailored views for five user groups and gives users agency over the explanations' content and their format. We evaluated IXAII through interviews with experts and lay users. Our results indicate that IXAII, which provides different explanations with multiple visualization options, is perceived as helpful to increase transparency. By bridging the gaps between explainable AI methods, interactivity, and practical implementation, we provide a novel perspective on AI explanation practices and human-AI interaction.

\end{abstract}
%
%
% NOTE:
% Submissions for the products & prototypes track should comprise the following sections in no more than 6 pages (excluding first page, references, etc., and formatted according to the Springer LNCS guidelines):
% Title page with information about the designers, Design of the artifact, Significance to research, Significance to practice, Evaluation of the artifact
%
%
%
% If available, screencast of or link to an implementation of the artifact
\section{Introduction}
Artificial intelligence~(AI) systems are increasingly used in high-stakes domains due to their high performance with deep learning~(DL) models, although those very systems are de facto black boxes for human users~\cite{HermEtAl2022}. % See p. 1
A resulting lack of user trust is one of the challenges of these systems and can be largely attributed to insufficient system transparency~\cite{Miller2019,PfeufferEtAl2023}. % See Miller p.1 and Pfeuffer p.1
Multiple scholars have examined the effect of explanations on users. % See p. 2
Their research indicates that their explanations can establish human trust~\cite{MartinezEtAl2023}. % See p. 38 
However, most of these explanation methods offer only static explanations and neglect the user perspective, reducing their effectiveness. We posit that user agency is a necessity for better user acceptance.

Human-AI interaction, situated at the intersection of AI and human-computer interaction~(HCI), emphasizes enhancing human involvement, allowing users to become co-creators in interactive AI systems. While interaction can improve decision making and trust, current static explanations limit users to data recipients rather than active participants. A human-centered approach is essential, placing users at the core of AI design~\cite{RaeesEtAl2024} allowing more understanding and giving them a sense of agency~\cite{MartinzeAndMaedche2023}. % See [1] and [3]

% Note 4 Paupi: Repo cloned in D:\!Uni\!INFORMATIK MSc\MA_Git\IXAII
Against this backdrop, we developed an interactive explainable intelligent system called IXAII\footnote{The implementation is available at \url{https://github.com/DiePaupi/IXAII}} (\underline{I}nteractive E\underline{x}plainable \underline{AI} \underline{I}nterface) with multiple types of explanations, using the four relevant~\cite{MartinezEtAl2023} XAI methods: LIME~\cite{LIME}, SHAP~\cite{SHAP}, Anchors~\cite{ANCHORS}, and DiCE~\cite{DICE}. Our prototype allows users to select one of five user perspectives as proposed by Barredo Arrieta et al.~\cite{ArrietaEtAl2020} (developer, user, business entity, regulatory entity, and affected party) and choose the explanations that best fit their needs and abilities, giving them greater control and supporting their understanding and trust of the employed machine learning~(ML) model. This, in turn, enables users to identify unwanted system behaviors, such as negative biases, and reflect upon the proposed decision.

%(e.g., problem statement, use cases, intended user groups, description of features)
\section{Design of the Artifact}

\subsection{Designing Explanations}

An explanation consists of two main aspects: its content (what is expressed) and its format (how it is presented). The most common format method for data explanations are \emph{tables}%, which constitute around $27\,\%$ of all visualizations used in textbooks
~\cite{MedleyRathEtAl2023}. Other widely used options are bar and column charts%as well as their stacked versions,
but naturally, there are a multitude of presentation possibilities. Among them, formal expressions~(e.g.,~Anchors' if-then-rules), models~(such as decision trees), or textual descriptions can also be used to present explanations, as they are comparatively easy to understand.
Similarly, the explanations' content can be grouped into different reference methods. Table~\ref{tab:reference_methods} shows an excerpt of several methods common in literature.
\vspace{-0.75cm}
\begin{table}[h]
	\centering \small
	\caption{Overview of reference methods.}
	\label{tab:reference_methods}
	\begin{tabular}{clc}
    \toprule
        \textbf{Method} & \textbf{Description} & \textbf{Sources}\\
    \midrule
	\midrule
		Inputs       & \parbox[l]{8.2cm}{Overview of all applied inputs}                                                    & \cite{LimEtAl2019}\\[0.07cm]
		\parbox[c]{2cm}{What Output}        & \parbox[l]{8.2cm}{Overview of the current and all other possible outputs}     & \cite{LimEtAl2019}\\
		%Certainty    & \parbox[l]{8cm}{Display of the system's (un)certainty about its current output}                    & \cite{LimEtAl2019,SilvaEtAl2023,MuellerEtAl2019}\\
	\midrule
		How          & \parbox[l]{8.2cm}{Information about the ML model's inner logic}                                      & \cite{LimEtAl2009,MuellerEtAl2019,WannerEtAl2022_social}\\[0.15cm]
		Why          & \parbox[l]{8.2cm}{Information about why the system derived the current output from the given inputs} & \parbox[c]{1.3cm}{$\,\,\,$\cite{ArrietaEtAl2020,HermEtAl2023,LimEtAl2009} \cite{LimEtAl2019,SilvaEtAl2023,MuellerEtAl2019}}\\[0.18cm]
	\midrule
		Why Not      & \parbox[l]{8.2cm}{Information about why another possible outcome was not derived w.r.t.~the derived one and the applied inputs} & \cite{LimEtAl2009,LimEtAl2019}\\[0.25cm]
		What If      & \parbox[l]{8.2cm}{Display of a simulated output based on altered inputs}                             & \cite{HermEtAl2023,LimEtAl2009,LimEtAl2019}\\[0.07cm]
		When         & \parbox[l]{8.2cm}{Overview of simulated inputs based on a desired output}                            & \cite{LimEtAl2009,LimEtAl2019,SilvaEtAl2023}\\
	\bottomrule 
	\end{tabular}
\end{table} 
\vspace{-0.5cm}

In terms of explanation design, the following recommendations exist: Explanations should to be contrastive and interactive~\cite{Miller2019,MuellerEtAl2019,WannerEtAl2022_social}. % See Miller p.3, Mueller p. 74, Wanner p. 33,43
Further, they should be of low to medium complexity~\cite{AliEtAl2023,HermEtAl2023,RamonEtAl2021} and give users a sense of agency~\cite{BertrandEtAl2023,BurtonEtAl2020}. % See Ali p. 33, Herm p. 11&12, Ramon p.20&21, and Bertrand p.13, Burton pp.223-224
They also need to be perceived as intuitive, understandable, and useful, but most importantly as trustworthy by users~\cite{WannerEtAl2022_social}. % See p. 41 & p. 43
Finally, explanations should only present relevant but sufficient information~\cite{Miller2019,MuellerEtAl2019}, which can be achieved by providing audience-specific explanations.

Barredo Arrieta et al.~\cite{ArrietaEtAl2020} identified five separate audiences: developers, business entities, regulatory entities, users, and affected third parties. Here, lay users and ML novices prefer local (example-based) explanations, for example~\emph{Why} or \emph{Why~Not} explanations~\cite{HermEtAl2023,GAMUT,RiberaAndLapedriza2019,WannerEtAl2022_social}. % See Herm p.11, GAMUT p.9, Ribera p.4, Wanner p.45
However, users with more ML expertise tend to employ global explanations more frequently, and expert users fluidly switch between both local and global explanations in their reasoning~\cite{GAMUT}. % See p. 9

\subsection{Design of IXAII}

\vspace{-0.5cm}
\begin{figure}
    \centering
    \includegraphics[width=1\textwidth]{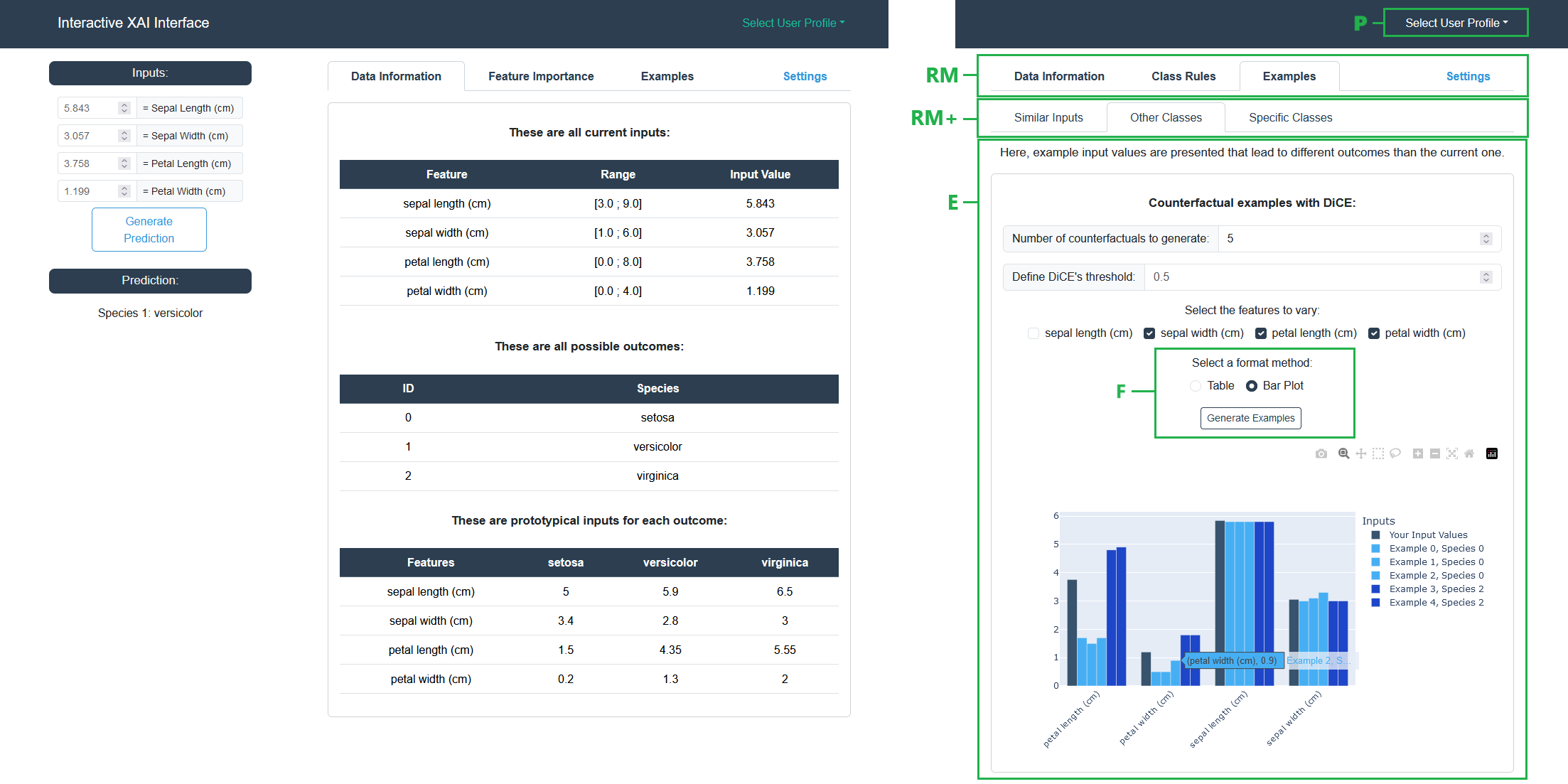}
    \caption{Screenshots of IXAII: Information about the dataset (left) and counterfactual examples %overlayed with the interface structure
  (right).}
    \label{fig:teaser}
\end{figure}
\vspace{-0.5cm}

As a first step, we derived nine design features~(DF) of such a system via a theory-motivated approach. Specifically, we conducted a hermeneutic literature review process~\cite{BoellAndCecez} to derive recommendations for generating explanations. When the literature review saturated, we formulated design principles~\cite{SpeckmannEtAl2025} and derived DFs that we further iterated in a user-centered design approach.
%TODO: Maybe just say that this is based on the full paper but already integrates some of the evaluation's feedback?
Notably, the version of IXAII presented here already incorporates improvements based on evaluation findings from~\cite{SpeckmannEtAl2025}.

The first three DFs are instantiated in the ``Data Information'' tab~(Figure~\ref{fig:teaser}, left) and address the reference methods' \emph{Inputs} and \emph{What~Output}. These methods focus on the underlying data and are independent of the system's predictions. Motivated by the idea that understanding the original training data can help explain an AI model's behavior~\cite{LimEtAl2019}, IXAII enables users to explore and examine the data through the instantiation of these DFs.

\begin{figure}[ht]
    \centering
    \includegraphics[width=0.98\textwidth]{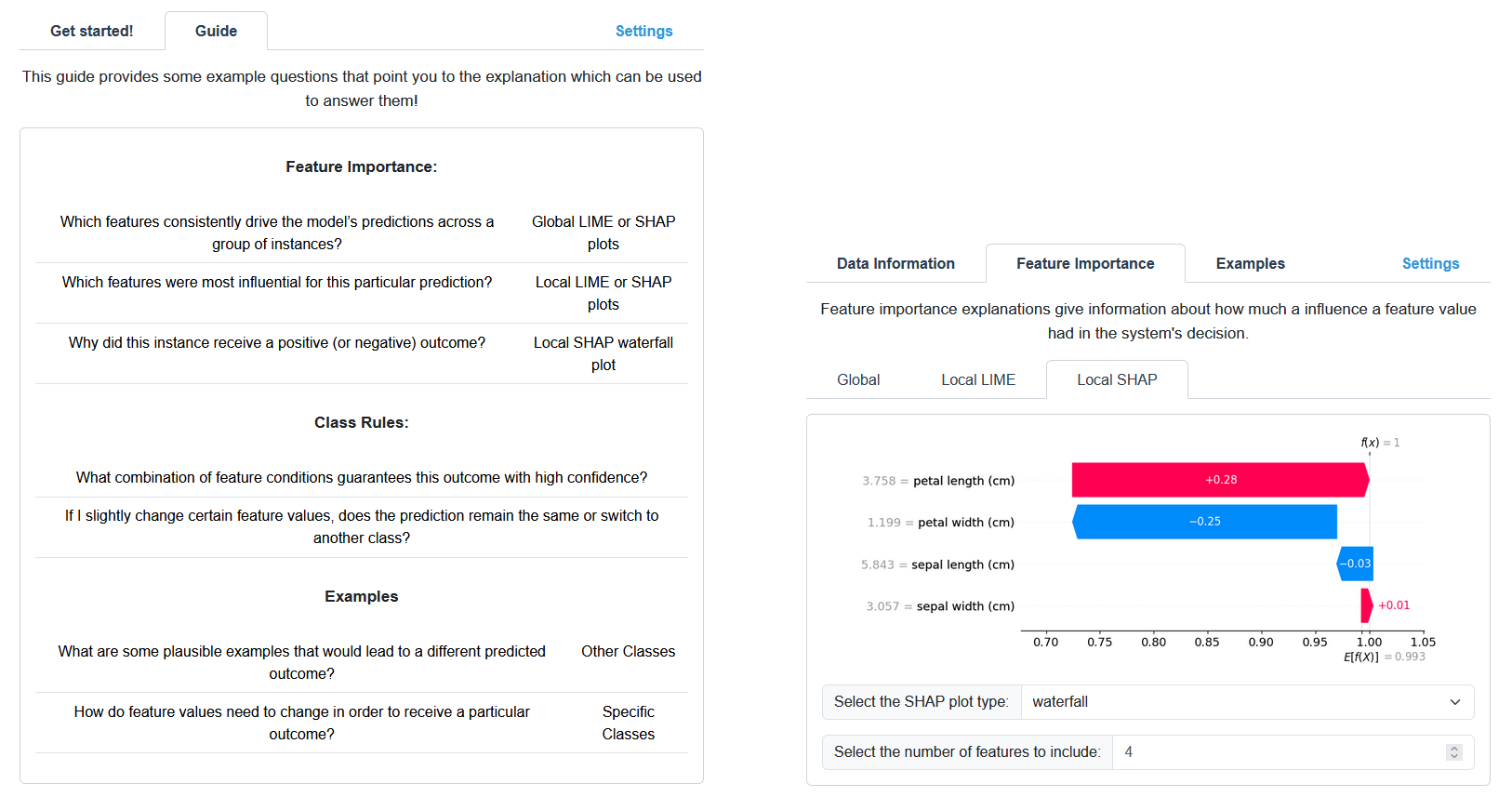}
    \caption{Screenshots of IXAII: Guide to XAI methods~(left) and local SHAP waterfall plot~(right).}
    \label{fig:IXAII}
\end{figure}

Furthermore, IXAII aims to offer users a diverse set of XAI and visualization tools. Users are able to select both the content and presentation style of explanations based on their questions and preferences~(DF$4$-$8$). They can switch between explanations by selecting the tabs displayed in areas~\textcolor{Green}{RM} and~\textcolor{Green}{RM$+$} of Figure~\ref{fig:teaser} (right) and their format in area~\textcolor{Green}{F}.
%\todo{farben sind schlecht zu lesen, dunkler wäre besser} 
The provided guide for the implemented XAI methods~(Figure~\ref{fig:IXAII}, left) helps especially lay users to quickly find explanations suitable to their questions.
%Opting for a simpler or more intuitive format reduces the complexity of the explanations, and allowing users to switch between different explanation approaches not only enhances their control, but also enables them to view multiple reasoning perspectives.
Giving users a choice about the displayed explanations enhances their control and enables them to view multiple reasoning perspectives.

Finally, by implementing DF$9$, IXAII addresses the importance of tailoring explanations to specific audiences~\cite{AliEtAl2023,ArrietaEtAl2020,HermEtAl2023,GAMUT,LarasatiEtAl2023,MuellerEtAl2019}.
By accommodating multiple target audiences (cf.~area~\textcolor{Green}{P} in Figure~\ref{fig:teaser}, right), the system ensures that each of the five user groups receives only relevant information~\cite{Miller2019}. %, since each of them has distinct information requirements~\cite{ArrietaEtAl2020}.
See Table~\ref{tab:design_features} for a list of all DFs considered.

\vspace{-0.75cm}
\begin{table}[h]
	\centering \small
	\caption{Considered DFs for IXAII (adapted from~\cite{SpeckmannEtAl2025}).}
	\label{tab:design_features}
	\begin{tabular}{cl}
	\toprule
		\textbf{DF} & \textbf{Description}\\
	\midrule
	\midrule
		1 & Provide an overview of all current inputs and their possible ranges\\
		2 & Provide an overview of all possible outcomes\\
		3 & Provide an overview of prototypical inputs for each outcome\\
	\midrule
		4 & Provide \emph{Certainty}, \emph{Why}, \emph{Why Not}, \emph{What If}, and \emph{When} reference methods\\
		5 & Provide the option to switch between reference methods\\
		6 & Provide a guide for the provided reference methods\\
	\midrule
		7 & Provide text, formal expression, table, chart, and explanation format methods\\
		8 & Provide the option to switch between format methods\\
		9 & Provide recommended views for each target audience\\
	\bottomrule 
	\end{tabular}
\end{table}
\vspace{-0.5cm}

%\vspace{-0.3cm}
%\vspace{-0.75cm}

For the implementation, we used Python, Scikit-learn~\cite{ScikitLearn}, and \href{https://dash.plotly.com/}{Plotly's Dash} for their seamless integration with various XAI methods and standardized interfaces.
%By using the \href{https://dash-bootstrap-components.opensource.faculty.ai/}{Dash Bootstrap Components} and its grid, we divided the interface into the separate sections as seen in Figure~\ref{fig:teaser}.
%
To focus on the interface rather than the data, we decided to use a minimalistic dataset, \href{https://scikit-learn.org/1.5/modules/generated/sklearn.datasets.load_iris.html}{IRIS}, with only four input features describing the width and length of the plant’s sepals and petals.
Similarly, the naming of the reference methods was found to be confusing, prompting us to choose more descriptive names for the explanations.

%(e.g., in laboratory or field settings)
% TODO: Yeet out the table & check quotes with the one from the full paper -> The data information tab is not seperated and available for everyone there yet...
\section{Evaluation of the Artifact}

For the evaluation of the IXAII prototype, we recruited seven participants: three experts~(E) and four lay users~(L). 
Each of the lay users is between $25$-$30$ years old and has between $0$-$1$ years of experience with ML, while each of the experts is between the ages of $30$-$35$, has $4$-$8$ years of experience with ML and holds a doctorate in information systems.
All interviews were conducted online and lasted between $60$ and $120\,$minutes, providing ample time for a comprehensive qualitative discussion. The participants received no payment for their contributions.

Once the participants had explored and used most, if not all, of the interface's features, they were asked to rate them. The overall impression of the interface was very positive. For instance, E1 mentioned that ``\emph{all key aspects of the topic [i.e.,~the interface] were covered}''. E2 found the interface helpful but noted that lay users would require more context to fully understand it. In this regard, L3 appreciated the interface, as they were able to answer most of the exercise questions independently despite having no prior knowledge. However, some critique was provided by E1 and E3, who both mentioned that developers would benefit from more data exploration options (e.g.,~more detailed histograms). %, though they acknowledged that such functionality would be beyond the scope of this project.
% - On a critical note, E3 commented that the interface currently felt more like a collection of individual explanation tools rather than a cohesive system.
%
Concluding the feedback on the improved IXAII version, L2 said that they found it ``\emph{cool, that [the data information tab] is accessible for all users}'' and L4 appreciated the guide and intuitive naming of the reference methods, as they ``\emph{make it very easy to find one's way around}''.

%(e.g., innovativeness of the artifact)
\section{Significance to Research} 

The development of IXAII provides significant contributions to research in XAI and HCI. By integrating multiple XAI methods such as SHAP, LIME, Anchors, and DiCE, IXAII addresses the diverse explanatory needs of different user groups. This adaptability is crucial, as research has yet to fully explore the varying requirements of stakeholders engaging with AI explanations. Our work not only enhances transparency in AI decision making but also lays the foundation for further studies on tailoring explanations to specific user needs.
A core theoretical contribution of IXAII lies in its emphasis on interactivity and user agency.
Interactivity has been shown to foster a sense of agency, which promotes trust as well as objective and subjective understanding~\cite{ChengEtAl2019,MartinzeAndMaedche2023,MuellerEtAl2019}, and aligns with perspectives suggesting that understanding emerges through recursive interaction rather than passive information absorption.
%IXAII supports a paradigm where explanations are not merely static outputs but dynamically co-constructed through user interaction. This theoretical positioning aligns with research on human cognition and underscores the importance of agency in fostering meaningful understanding of AI-driven decisions.

Beyond theoretical insights, IXAII also contributes to design research by deriving normative design knowledge on socio-technical AI explanations. By identifying and evaluating DFs that enhance human interpretability, our work adds to the growing body of knowledge on designing effective and user-centered AI systems. These insights can inform future research, ensuring that AI explanations remain comprehensible and actionable for diverse user groups.
%
%Overall, IXAII serves as both a research artifact and a methodological framework for studying interactive explanations in XAI. By bridging theoretical perspectives on agency and interactivity with empirical design knowledge, our work advances the discourse on how explainability can be meaningfully integrated into socio-technical AI systems.

%(e.g., usefulness of the artifact)
\section{Significance to Practice}

IXAII offers significant practical value by enabling users to engage with AI explanations in a way that best suits their needs. One key finding of our work is the importance of providing multiple explanation formats to allow users to choose the representation that aligns with their preferences and level of expertise. This flexibility enhances user comprehension and trust, as stakeholders can access explanations in a way that resonates with their cognitive and contextual requirements. By incorporating different explanatory approaches, IXAII ensures that AI transparency is not a one-size-fits-all solution but an adaptable framework that serves a diverse audience.

Furthermore, IXAII is designed to be versatile and applicable across various real-world scenarios. While there is still a long way to go, we provide a foundation where IXAII's interactive nature allows it to be adjusted for different domains, making it a valuable tool for research and industries where AI-driven decisions require explanation, such as healthcare, finance, and legal systems.
%Organizations can customize IXAII to provide stakeholders with explanations that align with their decision making processes. This adaptability ensures that AI explanations remain relevant and actionable across different contexts, promoting responsible AI deployment.
%
By offering both choice and flexibility in explainability, IXAII bridges the gap between AI developers and users, ensuring that explanations are not only technically sound but also practically meaningful. This positions IXAII as a valuable resource for fostering AI transparency in real-world applications while supporting further research into effective AI explanation strategies.

%%% Conclusion %%%%%%%%%%%%%%%%%%%%%%%%%%%%%%%%%%%%%%%%%%%%%%%%%%%%%%%%%%%%%%%%%%%%%%%%%%%%%%%%%%%%%%%
\section{Conclusion}
In this work, we developed the interactive explainable AI interface IXAII enabling the interactive use of a range of reference and format methods. The prototype serves as a practical demonstration of how interactivity and adaptability can be employed to enhance ML model explanations and enable agency in human-AI interaction.
%Drawing from existing XAI literature and user-centered goals, we formulated nine DFs normatively guiding the development of further prototypes. Our user feedback indicates that the interactive features not only improve user understanding but also enhance user satisfaction by enabling them to access explanations tailored to their tasks and preferences.
With our prototype, we demonstrate that user specificity is crucial as different users have varied goals and unique needs for explanations. 

%As the evaluation included only seven participants in total (three experts and four lay users), we plan to expand the sample size for future research and to include users from all target audiences to obtain a broader range of feedback and allow for a better generalization of preference differences between these user groups.

Our prototype highlights the value of designing explanations that not only present model insights but do so in an intuitive, interactive way, reducing the cognitive effort needed for understanding them and thus enhancing the user experience by introducing agency.

%In sum, we contribute to the field threefold: With IXAII, we introduce an interactive interface providing several explanations configured for different user groups. Second, we demonstrate its potential in problem-solving tasks by relying on an evaluation of experts and lay users. Third, we derive normative design knowledge in terms of design features that contribute to the growing body of knowledge on designing socio-technical AI explanations for human users.

%
% ---- Credits ----
%
\begin{credits}
    %\subsubsection{\ackname} A bold run-in heading in small font size at the end of the paper is
    %used for general acknowledgments

    \subsubsection{\discintname}
    The authors have no competing interests to declare that are relevant to the content of this article.
\end{credits}
\bibliographystyle{splncs04}
\bibliography{IXAII.bib}
\end{document}